\documentclass[11pt,a4paper]{article}
\usepackage{times,latexsym}
\usepackage{url}
\usepackage[T1]{fontenc}

\usepackage[acceptedWithA]{tacl2021v1}
\usepackage{tacl2021v1}
\usepackage{amsmath}
\usepackage{graphicx}
\usepackage{url} 

\usepackage{xspace,mfirstuc,tabulary}

\newif\iftaclinstructions
\taclinstructionsfalse 
\iftaclinstructions
\renewcommand{\confidential}{}
\renewcommand{\anonsubtext}{(No author info supplied here, for consistency with
TACL-submission anonymization requirements)}
\newcommand{\instr}
\fi

\iftaclpubformat 

\else

\fi

\title{How to Describe Images in a More Funny Way? Towards a Modular Approach to Cross-Modal Sarcasm Generation}

\author{
  Jie Ruan
  , \ 
  Yue Wu
  , \ 
  Xiaojun Wan
  , \ 
  Yuesheng Zhu
  \\
  Peking University
  \\
  \texttt{ruanjie@stu.pku.edu.cn}
  \\
  \texttt{zaczywy@pku.edu.cn}
  \\
  \texttt{wanxiaojun@pku.edu.cn}
  \\
  \texttt{zhuys@pku.edu.cn}
}

\date{}

\begin{document}
\maketitle

\begin{abstract}
Sarcasm generation has been investigated in previous studies by considering it as a text-to-text generation problem, i.e., generating a sarcastic sentence for an input sentence.  
In this paper, we study a new problem of cross-modal sarcasm generation (CMSG), i.e., generating a sarcastic description for a given image. CMSG is challenging as models need to satisfy the characteristics of sarcasm, as well as the correlation between different modalities. In addition, there should be some inconsistency between the two modalities, which requires imagination. Moreover, high-quality training data is insufficient.
To address these problems, we take a step toward generating sarcastic descriptions from images without paired training data and propose an Extraction-Generation-Ranking based Modular method (EGRM) for cross-model sarcasm generation. 
Specifically, EGRM first extracts diverse information from an image at different levels and uses the obtained image tags, sentimental descriptive caption, and commonsense-based consequence to generate candidate sarcastic texts. Then, a comprehensive ranking algorithm, which considers image-text relation, sarcasticness, and grammaticality, is proposed to select a final text from the candidate texts. 
Human evaluation at five criteria on a total of 1200 generated image-text pairs from eight systems and auxiliary automatic evaluation show the superiority of our method. 
\end{abstract}

\begin{figure}[]
\centering
\includegraphics[width=1\columnwidth]{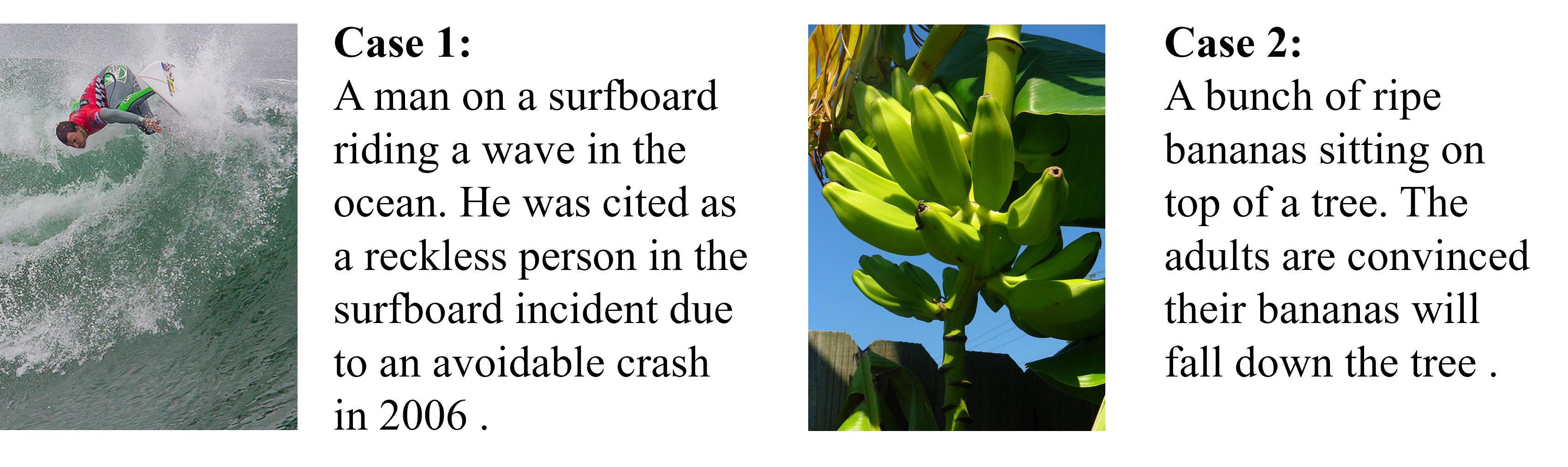} 
\caption{Generated examples of the cross-modal sarcasm generation task. Case 1 satirizes that the man may be crashed. Case 2 satirizes that the bunch of bananas is not yet ripe and adults always think they are right.}
\label{fig:intro}
\end{figure}

\section{Introduction}\label{sec:intro}
In Merriam Webster, sarcasm is defined as \emph{“a mode of satirical wit depending for its effect on bitter, caustic, and often ironic language that is usually directed against an individual”\footnote{https://www.merriam-webster.com/dictionary/sarcasm}.} 
The use of sarcasm is found to be beneficial for increasing creativity and humor on both the speakers and the addressees in conversations \cite{bowman2015large,burgers2012verbal}. Therefore, researches on sarcasm have an influence on downstream application tasks such as dialogue system, content creation, and recommendation.
Machines with sarcasm are often seen as intelligent, imaginative, and witty, which fits the key goal of \emph{Artificial general intelligence}\footnote{https://en.wikipedia.org/wiki/Artificial\_general\_intelligence}.

Over the years, studies have investigated sarcasm detection and textual sarcasm generation. Sarcasm detection aims to detect whether the input data is sarcastic, which has been explored in some research work \cite{davidov2010semi,gonzalez2011identifying,riloff2013sarcasm,joshi2015harnessing,ghosh2015sarcastic,muresan2016identification,ghosh2017magnets,ghosh2017role}. However, research on sarcasm generation stays in textual (text-to-text) sarcasm generation \cite{joshi2015sarcasmbot,peled2017sarcasm,zhu2019neural,mishra2019modular,chakrabarty2020,oprea2021chandler}, that is, outputting sarcastic text for the input text. Till now, there is no work attempting to generate sarcastic texts for images, while enabling machines to perceive visual information and generate sarcastic text will increase the richness and funniness of content or conversation, and serve downstream tasks such as content creation and dialogue systems. 
In this study, we for the first time formulate and investigate a new problem of cross-modal sarcasm generation (CMSG). 

Cross-modal sarcasm generation is a challenging problem as it should not only retain the characteristics of sarcasm but also make the information generated in a different modality related to the original modality.
In addition, there should be some inconsistency between the semantic information of the two modalities, which requires imagination. For example, the literal and intended meaning is reversed. 
The information of the two modalities should have the effect of enhancing or producing sarcasm.
Sarcasm factors are defined as follows: 1) be evaluative, 2) be based on the inconsistency of the ironic utterance with the context, 3) be based on a reversal of valence between the literal and intended meaning, 4) be aimed at some target, and 5) be relevant to the communicative situation in some way \cite{burgers2012verbal,burgers2011finding}. 
Moreover, there is insufficient high-quality cross-modal sarcasm training data, which makes cross-modal sarcasm generation more difficult. Experiment results on one of our baseline BLIP \cite{li2022blip} demonstrate that the existing cross-modal sarcasm dataset \cite{cai2019multi} is unable to solve problems in a supervised way.

\begin{figure*}[t]
\centering
\includegraphics[width=2\columnwidth]{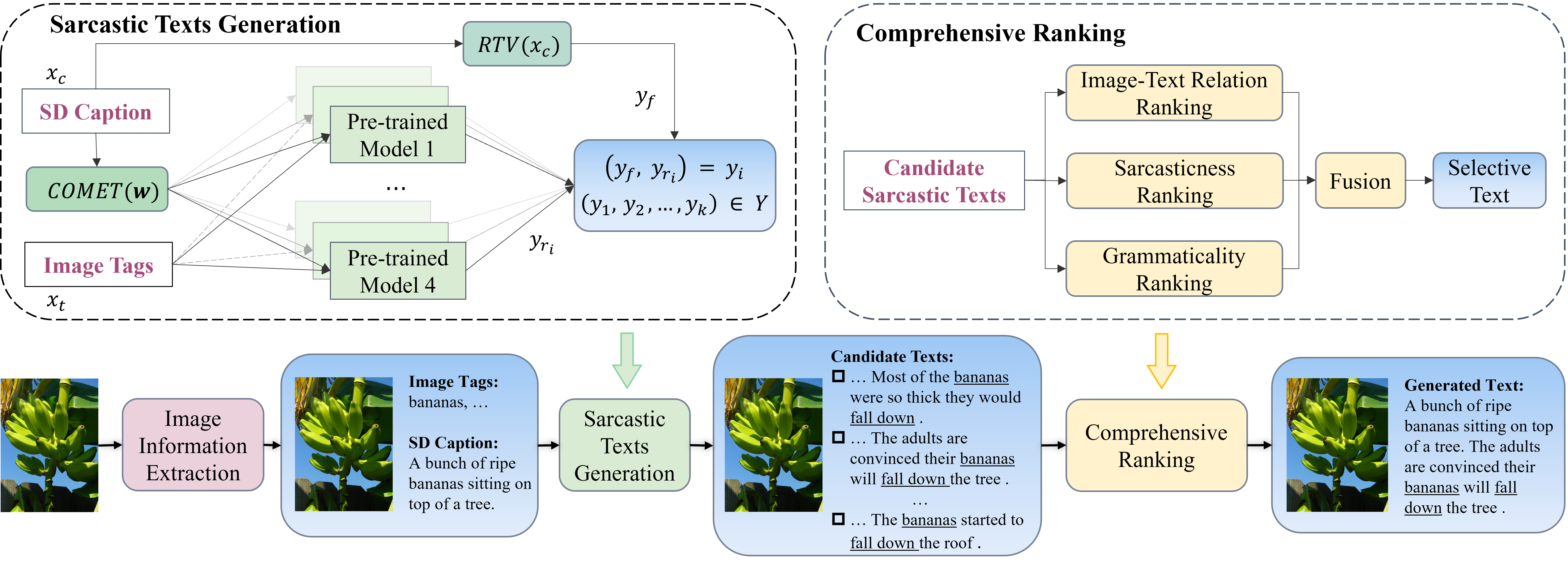} 
\caption{The overall framework of EGRM. EGRM consists of three modules: image information extraction, sarcastic texts generation, and comprehensive ranking. In the sarcastic texts generation module, RTV reverses the valence of the SD Caption. COMET is a commonsense reasoning method used to infer the consequence of the SD Caption. }
\label{fig:framework}
\end{figure*}

To address the above problems, we focus on generating sarcastic texts from images and propose an Extraction-Generation-Ranking based Modular method (EGRM) for unsupervised cross-modal sarcasm generation (shown in Figure \ref{fig:framework}).
We introduce to extract and obtain diverse image information at different levels through image tagging and sentimental descriptive captioning for generating sarcastic texts. A sarcastic texts generation module is proposed to generate a set of candidate sarcastic texts.
In the sarcastic texts generation module, we first reverse the valence (RTV) of the sentimental descriptive caption and use it as the first sentence. Then the cause relation of commonsense reasoning is adopted to deduce the consequence of the image information, and the consequence and image tags are used to generate a set of candidate sarcastic texts.
As the cross-modal sarcasm generation task involves the evaluation from multiple perspectives, we propose a comprehensive ranking method that considers image-text relation, sarcasticness, and grammaticality to rank the candidate texts. 
Two examples of the generated image-text pairs are shown in Figure \ref{fig:intro}.

The main contributions of our work are as follows: 1) For the first time, we formulate the problem of cross-modal sarcasm generation and analyze its challenges. 2) We propose a novel and non-trivial extraction-generation-ranking based modular method (EGRM) to address the challenging cross-modal sarcasm generation task. EGRM uses commonsense-based consequence and image tags to generate imaginative sarcastic texts, which makes the two modalities relevant and inconsistent to produce sarcasm.
Moreover, we consider the performance of candidate sarcastic texts from multiple perspectives, including image-text relation, semantic inconsistency, and grammar, and propose a comprehensive ranking method that simultaneously considers the performance of candidate texts from multiple perspectives to select the best-generated text. 
Our method does not rely on cross-modal sarcasm training data. 3) Human evaluation results show the superiority of our proposed method in terms of sarcasticness, humor, and overall performance. 

\section{Related Work}
\subsection{Textual Sarcasm Generation}
Research on Textual Sarcasm Generation is relatively preliminary. The limited amount of research on textual sarcasm generation is mainly divided into two categories, one is to generate a \emph{sarcasm response} based on the input utterance \cite{joshi2015sarcasmbot,oprea2021chandler}, and the other is to generate a \emph{sarcasm paraphrase} based on the input utterance \cite{peled2017sarcasm,mishra2019modular,chakrabarty2020}. 
Joshi et al. \shortcite{joshi2015sarcasmbot} introduced a rule-based sarcasm generation module named SarcasmBot. SarcasmBot implements eight rule-based sarcasm generators, each of which generates a kind of sarcasm expression. 
Peled and Reichart \shortcite{peled2017sarcasm} proposed a novel task of sarcasm interpretation which generate a non-sarcastic utterance conveying the same message as the original sarcastic utterance. They also proposed a supervised sarcasm interpretation algorithm based on machine translation. 
However, it is impractical to train supervised generative models with deep neural networks due to the lack of large amounts of high-quality cross-modal sarcasm data. Therefore, we turn to unsupervised approaches.
Mishra et al. \shortcite{mishra2019modular} introduced a retrieval-based framework that employs reinforced neural sequence-to-sequence learning and information retrieval and is trained only using unlabeled non-sarcastic and sarcastic opinions.
Chakrabarty et al. \shortcite{chakrabarty2020} presented a retrieve-and-edit-based framework to instantiate two major characteristics of sarcasm: reversal of valence and semantic incongruity with the context, which could include shared commonsense or world knowledge between the speaker and the listener.
Oprea1 et al. \shortcite{oprea2021chandler} proposed Chandler, a system not only generates sarcastic responses but also explanations for why each response is sarcastic.
However, these works are mainly generating sarcasm text based on input utterance, and there is no existing research on cross-modal sarcasm generation. Enabling machines to perceive visual information and generate sarcasm information for communication will increase the richness and humor of communication and serve downstream tasks such as content creation. Therefore, we focus on cross-modal sarcasm generation.
\subsection{Image Captioning}
Image Captioning is the task of describing the content of an image in words. 
Recent works on image captioning have concentrated on using the deep neural network to solve the MS-COCO Image Captioning Challenge\footnote{\url{http://mscoco.org/dataset/\#captions-challenge2015}}. 
CNN family is often used as the image encoder and the RNN family is used as the decoder to generate sentences \cite{vinyals2015show,karpathy2015deep,donahue2015long,yang2016review,wang2021high}. Many methods have been proposed to improve the performance of image captioning. Previous work used reinforcement learning methods \cite{ranzato2015sequence,rennie2017self,liu2018context}, high-level attributes detection \cite{wu2016value,you2016image,yao2017boosting}, visual attention mechanism \cite{xu2015show,lu2017knowing,pedersoli2017areas,anderson2018bottom,pan2020x}, contrastive or adversarial learning \cite{dai2017contrastive,dai2017towards}, scene graph detection \cite{yao2018exploring,yang2019auto,shi2020improving} , and transformer \cite{cornia2020meshed,li2019entangled,luo2021dual,ji2021improving,xian2022dual,mao2022rethinking,wang2022geometry,kumar2022dual}.
A slightly related branch of our research in image captioning is sentimental image captioning which generates captions with emotions. 
Mathews et al. \shortcite{mathews2016senticap} proposed SentiCap, a switching architecture with factual and sentimental caption paths, to generate sentimental descriptive captions. You et al. \shortcite{you2018image} introduced Direct Injection and Sentiment Flow to better solve the sentimental image captioning problem. Nezami et al. \shortcite{nezami2018senti} proposed an attention-based model namely SENTI-ATTEND to better add sentiments to image captions.
Li et al. \shortcite{li2021image} introduce an Inherent Sentiment Image Captioning (InSenti-Cap) method via an attention mechanism.
However, cross-modal sarcasm generation involves creativity as well as correlations and inconsistencies among different modalities, existing image captioning methods cannot meet the requirement.

\section{Methodology}
Due to the low quality and insufficient quantity of existing cross-modal sarcasm training data, which is confirmed in the experiment results of our pre-trained supervised baseline BLIP, we focus on unsupervised cross-modal sarcasm generation.
However, retrieval-based methods for generating sarcasm sentences are limited by the quality of the retrieval corpus and the ability of multi-keyword retrieval. The sentences generated by rule-based methods are easily limited by the proposed rules and have worse performance on tasks requiring creativity and imagination like sarcastic texts generation. Therefore, we propose a modular cross-modal sarcasm generation method, which has a key component of constrained text generation and is able to generate more imaginative and creative sarcastic texts.

The overall framework of our proposed Extraction-Generation-Ranking based Modular method (EGRM) is shown in Figure \ref{fig:framework}. Given an image, EGRM generates a sarcastic text related to the input image. EGRM consists of three modules: image information extraction, sarcastic texts generation, and comprehensive ranking, as shown in Figure \ref{fig:framework}. The image information extraction module extracts and obtains diverse image information at different levels, including image tags and sentimental descriptive caption (SD Caption). In the sarcastic texts generation module, we first reverse the valence (RTV) of the sentimental descriptive caption and use it as the first sentence. Then the cause relation of commonsense reasoning is adopted to deduce the consequence of the image information, and the consequence and image tags are used to generate a set of rest texts via constrained text generation techniques. The first sentence and each rest text are concatenated to form a candidate sarcastic text set. 
At last, we propose a comprehensive ranking module with multiple metrics to measure various aspects of the generated texts and the highly ranked one is selected. As shown in Figure \ref{fig:framework}, the candidate image-text pairs are ranked and selected by using multiple metrics.

\subsection{Image Information Extraction}
As a cross-modal sarcasm generation task, it is crucial to extract and obtain important and diverse information from the input image that is useful for generating sarcastic texts.
We obtain image tags $x_t$ and sentimental descriptive caption $x_c$ from the image. Particularly, a popular object detection method YOLOv5 \cite{glenn_jocher_2022_6222936} is adopted to detect objects in the image and record image tags. SentiCap \cite{mathews2016senticap}, a switching recurrent neural network with word-level regularization, is used to generate sentimental descriptive image caption. 



\subsection{Sarcastic Texts Generation}
As shown in the upper-left part of Figure \ref{fig:framework}, there are two branches in the sarcastic texts generation module.
The top branch generates the first sentence $y_f$ from the sentimental descriptive caption (SD Caption) $x_c$. The bottom branch generates a set of rest texts $(y_{r_1},y_{r_2},...y_{r_k})$ from the given sentimental descriptive caption $x_c$ and image tags $x_t$. $k$ represents the total number of generated texts. 
Concretely, we generate multiple rest texts by using different pre-trained models with different image tags and consequence collocations as input.
The first sentence is then concatenated with each generated rest text to produce a set of candidate sarcastic texts $Y$, where each candidate text $y_i \in Y$. 

The sarcastic texts generation method needs to satisfy the correlation between image and text and also the inconsistency of the two modalities.
This means that the content of the generated text should be related to the image. At the same time, there is some inconsistency in the semantic information of the generated text with regard to the image, such as forming inversion or obtaining some contrast content, which is related to the image but not directly reflected by the image, through certain imagination and reasoning.
Firstly, we obtain our first sentence $y_f$ based on the sentimental descriptive caption generated from the input image to achieve image-text relevance. We \textbf{r}everse \textbf{t}he \textbf{v}alence (RTV) of the caption to make the text and image inconsistent.
Considering that sarcasm usually occurs in positive sentiment towards a negative situation (i.e., sarcastic criticism) \cite{chakrabarty2020,kreuz2002asymmetries}, we invert the negative sentiment expressed by the caption, so that the first sentence contains context with positive sentiment. 
Specifically, we obtain the negative score of the evaluative word from SentiWordNet \cite{esuli2006sentiwordnet} and use WordNet \cite{miller1995wordnet} to replace the evaluative words with its antonyms similar to the $R^3$ method \cite{chakrabarty2020}. We do nothing if there is no negative sentiment in the caption.
To sum up, the first sentence is obtained as $y_f = \text{RTV}(x_c)$. For example, for a raining image, we may reverse the first sentence ``a \textbf{bad} rainy day'' to ``a \textbf{good} rainy day'', which produces sarcasm and humor and may enhance sarcasm by the rest generated text.

The key to producing sarcasm is the reversal of valence between the literal and intended meaning as well as the relevance of the communicative situation. In the CMSG task, we should make some semantic inconsistency between the connotation expressed by the text and the real information shown by the image in the specific situation of the image.
To achieve this goal, we propose to use the commonsense-based consequence inferred by information from the image modality and the image tags to generate the rest texts, which will be concatenated after the first sentence. 
The reason we use the image information to deduce the consequence $c$ is that commonsense reasoning can infer the cause relation and the possible consequence in the scene shown in the image, making the intention of the sarcasm clearer and the effect of the sarcasm more intense. 
Taking the first example in Figure \ref{fig:intro} as an instance, commonsense reasoning result shows that information in the image may cause a \textbf{crash}. We may not feel sarcastic when we read the first sentence ``a man on a surfboard riding a wave in the ocean''. However, we feel sarcastic and funny when we imagine a man riding a wave and suddenly falls down from the surfboard which causes a crash. By using the commonsense-based consequence, the model is able to capture the deeper information contained in the image and imagine possible situations based on the commonsense-based consequence to generate more realistic sarcastic texts.
For inferring commonsense-based consequence, we extract verbs, nouns, adverbs, and adjectives, which denote as $\mathbf{w}$, from the sentimental descriptive caption $x_c$ and feed them to COMET to infer the consequence. Detailed information can be seen in these papers \cite{bosselut2019comet,chakrabarty2020,speer2017conceptnet}. Therefore, the commonsense-based consequence $c$ is obtained by $c = \text{COMET}(\mathbf{w})$.

Using image tags makes the image and text more relevant and makes it clearer who caused the consequence. In this way, we can generate sarcastic texts that are related to the image and inconsistent with the real semantic content.
For instance, both SC-$R^3$ and our method infer the consequence ``crash'' of the first example in Figure \ref{fig:intro}. SC-$R^3$ retrieves sentences from the corpus according to the commonsense-based consequence and gets a sentence ``The ceiling came down with a terrific crash.'', which is irrelevant to the image. The result is not only non-ironic but also confusing. Our method considers image tags and the commonsense-based consequence, and the generated texts have image-text correlation and inconsistency, which makes this image-text pair produce sarcasm.

To implement the cross-modal sarcastic texts generation module, we generate the rest texts based on a recently proposed constrained text generation method CBART \cite{he2021parallel}.
For instance, given image tag ``bananas'' and consequence ``fall down'' as input, the model may generate ``The adults are convinced their bananas will fall down the tree'', which can be seen in Figure \ref{fig:framework}.
As shown in the upper-left part of Figure \ref{fig:framework}, by using different numbers of tags, changing different pre-trained models, and using commonsense-based consequence inferred by information from the image modality, the sarcastic texts generation module can generate a variety of different sarcastic texts for selection. We use four pre-trained models to generate texts which are the base model initialized with BART-base model training on One-Billion-Word \cite{chelba2013one} dataset (base-One-Billion-Word), the base model initialized with BART-base model training on Yelp\footnote{https://www.yelp.com/dataset} dataset (base-Yelp), the large model initialized with BART-large model training on One-Billion-Word dataset (large-One-Billion-Word), and the large model initialized with BART-large model training on Yelp dataset (large-Yelp). Different pre-trained models can generate 
diverse rest texts, making our candidate sarcastic texts more abundant.
For more details about CBART, please read the paper of CBART \cite{he2021parallel}.

\subsection{Comprehensive Ranking}
In the CMSG task, we need to convert the image to the text of the target sarcasm style $s_t$.
Given an input image $x$, the conditional likelihood of the generated sarcastic text $y$ is divided into three terms:
\begin{equation}
\begin{aligned}
    p(y \mid x, s_{t})
    &=\frac{p(y,x, s_{t})}{p(x, s_{t})}
    \propto p(x,[y, s_{t}])\\
    &=\ p(x \mid[y, s_{t}]) \ \  p([y, s_{t}])\\
    &=\underbrace{p(x \mid [y, s_{t}])}_{\text {Image-Text Relation }} 
    \underbrace{p(s_{t} \mid y)}_{\text {Sarcasticness}} \underbrace{p\ ({\ y \ })}_{\text {Grammaticality}}, 
\end{aligned}\label{eq:rank}
\end{equation}
where $[\cdot]$ groups related terms (e.g., $[y, s_{t}]$) together. In the CMSG task, the first term of Equation \ref{eq:rank}, $p(x\mid[y, s_{t}])$ measures the \emph{Image-Text Relation} between the input image $x$ and the output target text $y$. It 
calculates the correlation between the image and the generated text. The second term, $p(s_{t} \mid y)$, can be seen as a measure of \emph{Sarcasticness}. The third term, $p(y)$, measures the overall \emph{Grammaticality} of the output text $y$, which also shows the fluency of the generated text.

Finally, we rank our $k$ candidate sarcastic texts generated in the cross-modal sarcastic texts generation module according to the decomposition in Equation \ref{eq:rank}. For the $i$-th candidate text $y_i$, the ranking score is computed as:
\begin{equation}
\begin{aligned}
    p_{crank}(y_i \mid x, s_{t}) \propto p(x \mid[y_i, s_{t}]) \ p\ (s_{t} \mid y_i) \  p\ (y_i), 
\end{aligned}\label{eq:crank}
\end{equation}
where $p_{crank}$ represents the comprehensive ranking probability for $y_i$. 
We choose the size of candidate sarcastic texts $k$ by conducting experiments on the validation data. Finally, the average $k$ of our method is 36.

All that remains is to describe how to calculate each term in Equation \ref{eq:crank}. To calculate the first term, image-text relation, we adopt a reference-free metric CLIPScore \cite{hessel2021clipscore} which measures the cosine similarity between the visual CLIP \cite{radford2021learning} embedding $v$ of the image $x$ and the textual CLIP embedding $e$ of a candidate text $y_i$. We presume $p(x\mid[y_i, s_{t}]) = \text{CLIPScore}(x,y_i) = w \cdot max(cos(e, v), 0)$ and $w$ follows the settings of CLIPScore \cite{hessel2021clipscore}, which is set as 2.5.
For calculating the second term, sarcasticness, we use semantic incongruity ranking \cite{chakrabarty2020} which fine-tunes RoBERTa-large \cite{liu2019roberta} on the Multi-NLI \cite{skalicky2018linguistic} dataset to calculate the contradictory score between the first sentence of the image description after reversing the valence and the rest text.
For the third term, we use perplexity (PPL) to calculate the existing probability of the texts, and we use the pre-trained model BERT \cite{kenton2019bert} to calculate the probability.

\section{Experimental Setup}
\subsection{Dataset}
As we do not need parallel cross-modal sarcasm data for training, we conduct the experiment on a testing subset of 503 images in the SentiCap \cite{mathews2016senticap} dataset, which uses images from the MSCOCO \cite{lin2014microsoft} validation partition and adds sentiment captions to those images.  
Automatic metrics for each method are calculated on these 503 images. Considering the time and economic cost of human evaluation, we randomly selected 150 images as the test set for human evaluation. Since there are eight systems, the human evaluation is conducted on a total of 1200 image-text pairs.

Datasets for training pre-trained models for the sarcastic texts generation module are the One-Billion-Word \cite{chelba2013one} dataset and the Yelp\footnote{https://www.yelp.com/dataset} dataset.
One-Billion-Word is a public dataset for language modeling produced from the WMT 2011 News Crawl data. The Yelp dataset contains business reviews on Yelp.

\begin{table*}[]
\centering
\resizebox{\textwidth}{2.1cm}{
\begin{tabular}{|c|c|c|c|c|c|c|c|}
\hline
\textbf{Method}                          & \textbf{TL}                  & \textbf{CLIPScore}           & \textbf{Sarcasticness}      & \textbf{Image-Text Relation} & \textbf{Humor}              & \textbf{Grammaticality}     & \textbf{Overall}            \\ \hline
\textbf{SC-MTS}\shortcite{mishra2019modular}                          & 9.43                         & 19.70                        & 0.65                        & 0.98                         & 0.71                        & 0.88                        & 0.73                        \\ \hline
\textbf{BLIP}\shortcite{li2022blip}                            & 9.87                         & \textbf{27.23}               & 1.31                        & \textbf{3.29}                & 1.91                        & 3.31*                       & 1.95                        \\ \hline
\textbf{SC-$R^3$}\shortcite{chakrabarty2020}                           & 19.11*                       & 25.15                        & 2.22*                       & 2.86                         & 2.21*                       & 3.30                        & 2.29*                       \\ \hline
\textbf{EGRM} (Ours)                             & \textbf{25.65}               & 25.31*                       & \textbf{2.85}               & \textbf{3.29}                & \textbf{2.78}               & \textbf{3.41}               & \textbf{2.90}               \\ \hline \hline
{\textbf{EGRM-woCS}} & {24.99} & {25.14} & {2.24} & {2.97}  & {2.27} & {3.37} & {2.38} \\ \hline
\textbf{EGRM-woTag}                       & 25.99                        & 24.78                        & 2.26                        & 2.91                         & 2.28                        & 3.32                        & 2.37                        \\ \hline
\textbf{EGRM-woS}                         & 30.99                        & 24.12                        & 2.39                        & 2.91                         & 2.33                        & 3.16                        & 2.42                        \\ \hline
\textbf{EGRM-woGI}                        & 26.24                        & 25.25                        & 2.34                        & 2.90                         & 2.28                        & 3.18                        & 2.39                        \\ \hline
\end{tabular}}
\caption{Evaluation results of all methods. The scores in columns 4$\sim$8 are human evaluation results, and the scale ranges from 0 (not at all) to 5 (very). The upper part of the table shows the comparison of our method and three baseline methods, and the lower part shows the results of our ablation study. As shown in the upper part of the table, our proposed EGRM has the best performance among all methods on all metrics except CLIPScore, on which EGRM is ranked 2nd (denoted by *).}
\label{tab:res}
\end{table*}
\subsection{Compared Methods}
As CMSG is a new task, we design the following three comparison methods, and the first two methods do not need parallel cross-modal sarcasm training data while the third one relies on such data for training. 
\begin{itemize}
     \item \textbf{SC-$R^3$}: We use the $R^3$ model released by Chakrabarty et al. \cite{chakrabarty2020} as it is the state-of-the-art textual sarcasm generation system to transform input texts into sarcastic paraphrases. We input the captions generated by SentiCap \cite{mathews2016senticap} to $R^3$ to generate sarcastic texts.
     \item \textbf{SC-MTS}: We input the captions generated by SentiCap to MTS \cite{mishra2019modular} to generate sarcastic texts. 
     \item \textbf{BLIP}: This is a pre-trained image captioning model \cite{li2022blip}, and we fine-tune it on the parallel cross-modal sarcasm dataset proposed by Cai et al. \cite{cai2019multi}. It is considered a representative of the supervised methods.
\end{itemize}
To explore the effectiveness of various parts of our proposed model EGRM, we ablate some components of EGRM and evaluate their performance. These are termed as: 
\begin{itemize}
    \item \textbf{EGRM-woCS}: the EGRM method without using the commonsense-base consequence to generate sarcastic texts. The goal of EGRM-woCS is to analyze the effect of commonsense reasoning consequence in the sarcastic texts generation module.
    \item \textbf{EGRM-woTag}: the EGRM method without using image tags to generate sarcastic texts. The goal of EGRM-woTag is to analyze the effect of image tags in the sarcastic texts generation module.
    \item \textbf{EGRM-woS}: the EGRM method without using sarcasticness ranking during comprehensive ranking. The goal of EGRM-woS is to analyze the effect of sarcasticness ranking in the comprehensive ranking module.
    \item \textbf{EGRM-woGI}: the EGRM method without using grammaticality ranking and image-text relation ranking during comprehensive ranking. The goal of EGRM-woGI is to analyze the effect of grammaticality ranking and image-text relation ranking in the comprehensive ranking module.
    \item \textbf{EGRM}: the complete method with all components.
\end{itemize}

\subsection{Evaluation Criteria}
The difficulty of evaluating the CMSG task is that it is a creative and imaginative study, and there is no standard sarcastic text for reference. In addition, the difference in the average text length generated by different methods may cause problems in traditional generation evaluation metrics. These reasons make traditional generation evaluation metrics like BLEU \cite{papineni2002bleu}, one of the most popular evaluation metrics in text generation tasks, unsuitable in the CMSG task involving creativity and imagination. This problem also exists in textual sarcasm generation task \cite{mishra2019modular,chakrabarty2020}. Therefore, human evaluation is mainly used for evaluation, and we use ClipScore \cite{hessel2021clipscore}, a popular reference-free image captioning metric, to evaluate the image-text relevance. 
Referring to the textual sarcasm generation metric WL \cite{mishra2019modular} for calculating the percentage of length increment, the notion behind which is that sarcasm typically requires more context than its literal version and requires to have more words present at the target side, we calculate the length of the generated text to assist in evaluating the performance of the model, and we name this metric total length (TL).
For human evaluation, we evaluate a total of 1200 generated image-text pairs since there are eight different systems with 150 image-text pairs each in our research. 

\begin{figure}[t]
\includegraphics[width=0.45\textwidth]{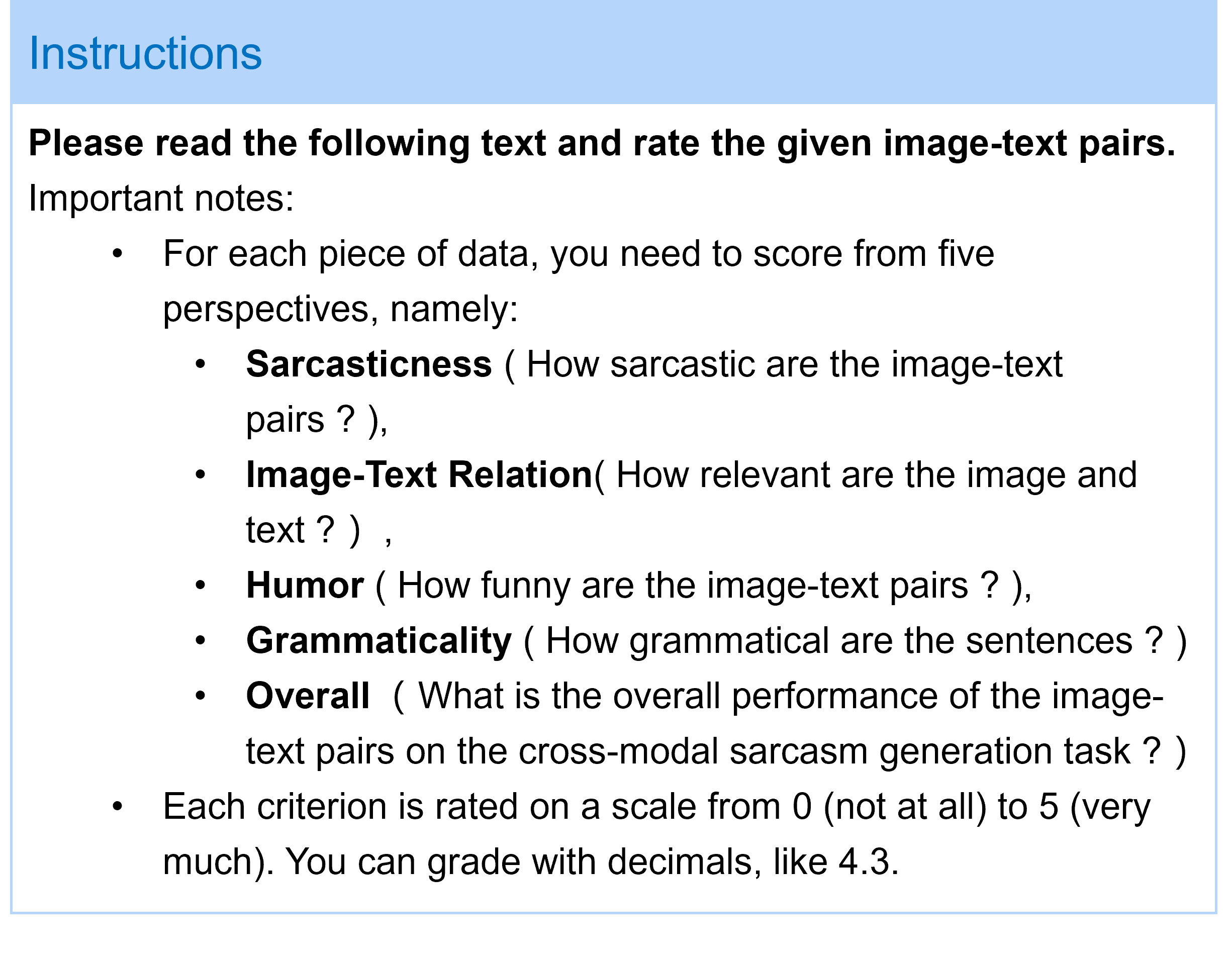} 
\caption{Instructions for human evaluation.}
\label{fig:instruction}
\end{figure}

Inspired by the evaluation method in previous work \cite{chakrabarty2020}, we propose five criteria to evaluate the performance of the cross-modal sarcasm generation methods: 1) \textbf{Sarcasticness} (How sarcastic is the image-text pair?), 
2) \textbf{Image-Text Relation} (How relevant are the image and text?), 
3) \textbf{Humor} (How funny is the image-text pair?) \cite{skalicky2018linguistic}, 
4) \textbf{Grammaticality} (How grammatical are the texts?) \cite{chakrabarty2020}, 
5) \textbf{Overall} (What is the overall quality of the image-text pair on the cross-modal sarcasm generation task?). 
We design an MTurk CMSG task where each Turker was asked to score the image-text pairs from all the eight methods. Each Turker was given the image together with a set of sarcastic texts generated by all eight systems. Each criterion is rated on a scale from 0 (not at all) to 5 (very much). The Turker can grade with decimals like 4.3. As CMSG is a difficult task requiring imagination, each image-text pair was scored by three individual Turkers. Each Turker is paid \$281.08 for the whole evaluation of 1,200 image-text pairs, which is roughly \$0.23 per image-text pair. Figure \ref{fig:instruction} shows the instructions released to the Turkers.

\section{Experimental Results}
\begin{figure*}[t]
\includegraphics[width=2.1\columnwidth]{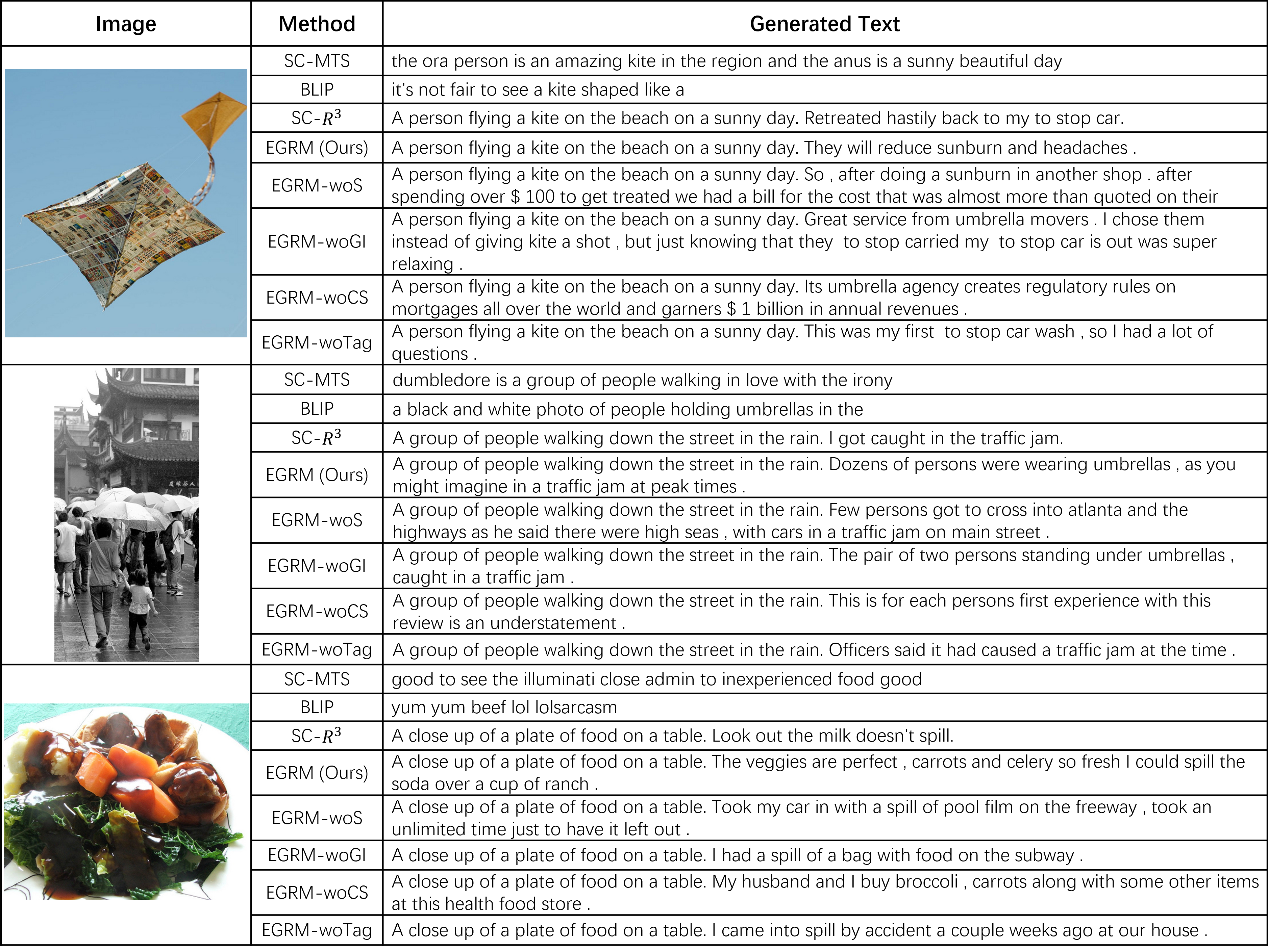} 
\caption{Examples of generated outputs from different systems.}
\label{fig:res-example}
\end{figure*}

\subsection{Quantitative Results}
Table \ref{tab:res} shows the scores on automatic metrics and human evaluation metrics of different methods. As shown in the upper part of the table, our proposed EGRM has the best performance among all comparison methods on all metrics except CLIPScore, on which EGRM ranks second.
The ablation study in Table \ref{tab:res} demonstrates that our full model EGRM is superior to ablation methods in all criteria except the total length. 
In terms of sarcasticness, our full model attains the highest average score, which shows our model meets the most important requirement of the CMSG task. According to the scores, EGRM gets the highest score on the humor criteria, which shows the potential contribution of our method for improving the interestingness and humor in content creation and communication. Moreover, the grammaticality of EGRM is good and the overall score of EGRM is the highest among all the methods.
The total length of the generated paragraph of EGRM is longer than SC-MTS, BLIP, and SC-$R^3$. This can be seen as an auxiliary basis for sarcasm as sarcasm typically requires more context than its literal version and requires to have more words present on the target side.

On the CLIPScore metric, we observe that EGRM does not have better performance than the pre-trained image captioning method BLIP, which is designed for generating textual descriptions of images. However, the CMSG task requires imagination and the method should imagine and generate text that is inconsistent with the image as well as relevant to the image, which leads to the CLIPScore of our method designed for the CMSG task being no better than the pre-trained image captioning method BLIP. Moreover, we can observe that EGRM and BLIP have the best performance among all the four methods on the image-text relation criteria in human evaluation. This is because when human judges consider whether the text is related to the image, they may allow reasonable imagination. Although BLIP has a higher CLIPScore, it cannot solve the CMSG problem due to the poor performance on sarcasticness. 
This also shows the existing parallel cross-modal sarcasm data is unable to train a good supervised model for the CMSG task, due to the limitations in scale and quality. 

\subsection{Ablation Study}
We concentrate our ablation study on the criteria of sarcasticness and overall performance, as we consider these metrics as the main criteria for the success of cross-modal sarcasm generation. 
As shown in Table \ref{tab:res}, the full model (EGRM) outperforms the other four ablation methods. 

EGRM-woCS has the worst performance in terms of sarcasticness among the ablation methods. This indicates that the commonsense-based consequence used in the sarcastic texts generation module, which is the inferring result of the image information based on commonsense reasoning, is important for sarcasticness. This is because the inconsistency between the commonsense reasoning consequence and the information of the image modality is the key to generating sarcasticness.
EGRM-woTag has the worst overall performance among the ablation methods. Because the combination of image information and inferring consequence can generate sarcastic image-text pairs where the two modalities are relevant, a text that is not related to the image may be regarded as incomprehensible in the generated text.
The experimental results of EGRM-woCS and EGRM-woTag show that the use of image tags and commonsense-based consequences in the generation module is crucial to generating image-text related and imaginary sarcastic texts.

EGRM-woS ranks first among the four ablation methods in terms of sarcasticness and overall performance while EGRM-woGI is slightly worse than EGRM-woS. However, both EGRM-woS and EGRM-woGI are worse than EGRM with a large margin, which demonstrates the importance of the three ranking criteria.
Moreover, image-text relation criteria are significant for sarcasticness because sarcasm is based on the correlation between the text and the image. If the text is not related to the image, the sarcasm is more likely to be poor, and sometimes it will be incomprehensible.

\subsection{Qualitative Analysis}
Figure \ref{fig:res-example} demonstrates several examples generated from different methods. 
Taking the text generated by EGRM from the first image in Figure \ref{fig:res-example} as an example, the image shows a kite flying in the sun. The person flying the kite is more likely to be full of joy. However, they may be kite flyers who may suffer from sunburn from overexposure to the sun and headaches from heat stroke. The pleasure of the image modality and the pain of the sunburn and the headache in the text modality are inconsistent, which produces sarcasm.
Moreover, the kite-flyers may think that the kite can help them block the sun and reduce sunburn and headaches, which is sarcastic about the stupidity of the kite-flyers. 
However, the results of SC-MTS and BLIP seem not to be sarcastic and the result of SC-$R^3$ seems to be confusing.
The second example shows that our approach is imaginative and humorous. EGRM imagines many people wearing umbrellas as traffic jams, and it satirizes road congestion caused by lots of umbrellas. The sarcasticness and humor score of EGRM in the second example is 4.33 and 3.83.
The third image shows a plate of food that does not look delicious. However, EGRM says that the veggies are perfect and the carrots are fresh, which makes the deliciousness displayed in the text and the bad taste displayed in the image reversed and inconsistent, making the image-text pair sarcastic. The other three comparison methods do not seem to produce sarcasm.

\section{Conclusion and Future Work}
We are the first to formulate the problem of cross-modal sarcasm generation and analyze the challenges of this task. We focus on generating sarcastic texts from images and proposed an extraction-generation-ranking based modular method with three modules to solve the problem without relying on any cross-modal sarcasm training data. Quantitative results and qualitative analysis reveal the superiority of our method.

In future work, we will explore generating sarcasm of different styles or categories. We will also try to build a large-scale high-quality parallel cross-modal sarcasm dataset for future researches in this field.

\bibliography{tacl2021}
\bibliographystyle{acl_natbib}








  

\end{document}